\documentclass[11pt,a4paper]{article}
\usepackage[hyperref]{eacl2021}
\usepackage{times}
\usepackage{latexsym}

\usepackage[ethiop,main=english]{babel}
\usepackage{tipa}

\usepackage{microtype}

\aclfinalcopy 
\def\aclpaperid{20} 

\title{Manually Annotated Spelling Error Corpus for Amharic}

\author{Andargachew Mekonnen Gezmu \\
Otto von Guericke Universität \\
Magdeburg, Germany \\
\texttt{andargachew.gezmu@ovgu.de} \\
\\
\textbf{Tirufat Tesifaye Lema} \\
Hawassa University \\
Hawassa, Ethiopia \\
\texttt{tirufatt@hu.edu.et} \\
\And
Binyam Ephrem Seyoum \\
Addis Ababa University \\
Addis Ababa, Ethiopia \\
\texttt{binyam.ephrem@aau.edu.et} \\
\\
\textbf{Andreas Nürnberger} \\
Otto von Guericke Universität \\
Magdeburg, Germany \\
\texttt{andreas.nuernberger@ovgu.de}
}

\date{}

\begin{document}
\maketitle
\begin{abstract}
This paper presents a manually annotated spelling error corpus for Amharic, lingua franca in Ethiopia. The corpus\footnote{Available at: \url{https://github.com/andmek/ErrorCorpus}} is designed to be used for the evaluation of spelling error detection and correction. The misspellings are tagged as non-word and real-word errors. In addition, the contextual information available in the corpus makes it useful in dealing with both types of spelling errors.
\end{abstract}

\section{Introduction}

Amharic, the official language of Ethiopia, is the second-most widely spoken Semitic language next to Arabic. It is also spoken by hundred thousands of Ethiopian immigrants in North America and Ethiopian Jews in Israel. In spite of its great number of speakers and being a written language for more than five
centuries, Amharic is still a scarce resource language as far as computational resources are concerned (\citealp{gezmu-etal-2018-portable}; \citealp{tracey-strassel-2020-basic}).

For researchers who are concerned for the development of Amharic spelling error correctors,  there is no standard error corpus to evaluate their approach. Spelling error corpora can be collected automatically from keystroke logs or word-typing games.  \citet{baba-suzuki-2012-spelling} extracted pairs of  misspellings and corrections from input logs by using Amazon’s Mechanical Turk. Researchers also attempt to collect such corpora from word-typing games (\citealp{rodrigues2012typing}; \citealp{tachibana2016analysis}). However, the former study is limited to languages supported by the crowdsourcing, and in the latter studies, subjects may behave differently than writing a regular text. Thus, we have developed a manually annotated spelling error corpus for Amharic.

\section{Related Work}

\citet{grudin1983error} offers a detailed study of the general pattern of errors made by novice and expert typists, where the subjects are asked to transcribe a text without correcting the errors they made. The researcher compiled letter confusion matrices in which typographical errors are categorized according to the letter intended and the letter actually struck. Even though it might be used in analyzing and modeling sources of misspellings, its lack of contextual information limits its scope of usage
particularly for real-word errors. A manually tagged spelling error corpus with contextual information is available from the book ``English for the Rejected'' in the Oxford Text Archive \citep{mitton1985collection}. Despite it was originally handwritten by poor spellers, its contextual information makes it still useful for evaluation purposes. Furthermore, \citealp{mitton2010fifty} argues that misspellings produced by poor spellers are worse than those caused by typing slips; a spelling corrector that can deal with poor spellings has a good chance of handling typos, but the reverse might not be true.

\section{Types of Spelling Errors}

The spelling errors in Amharic can be grouped as non-word and real-word errors. When typographical or cognitive errors accidentally produce valid Amharic words we get real-word errors, otherwise, we get  non-word errors. Typographical errors include insertion, deletion, transposition, and substitution of letters. Missed out spaces are also sources of typos.

The cognitive errors in Amharic mainly result from the inconsistency of its writing system, Ethiopic. Though Ethiopic shares most feature of abugida, it is considered as syllabary. Amharic has 27 consonant phonemes and seven vowels. Four of these phonemes have one or more homophonic character representations. The homophonic characters are the source of many cognates -- some scholars consider them as homonyms -- (\emph{e.g.}, \foreignlanguage{ethiop}{.sahaye}, \foreignlanguage{ethiop}{.cahaye}, \foreignlanguage{ethiop}{.sahAye}, \foreignlanguage{ethiop}{.sa.haye}, \foreignlanguage{ethiop}{.sa.hAye}, \foreignlanguage{ethiop}{.cahAye}, \foreignlanguage{ethiop}{.ca.haye} and \foreignlanguage{ethiop}{.ca.hAye} pronounced as  /{s'\textipa{@}haj}/ meaning  ``sun''). The general practice for strict Amharic writing style is that spellings of Amharic words inherited from Ge’ez, a parent language of Amharic, should follow Ge’ez features as much as possible, and loan words that use homophonic characters should be written only with \foreignlanguage{ethiop}{ha} /{ha}/, \foreignlanguage{ethiop}{sa} /{s\textipa{@}}/, \foreignlanguage{ethiop}{.sa} /{s'\textipa{@}}/ and \foreignlanguage{ethiop}{'a} /{\textipa{P}a}/ not with their variants \citep{cowley1967standardisation}. As such, real-word errors might occur from wrongly typed homonyms. For example, \foreignlanguage{ethiop}{se'ele} /{s\textipa{1}’\textipa{1}l}/ is a real-word error for \foreignlanguage{ethiop}{'se`ele} /{s\textipa{1}’\textipa{1}l}/ meaning ``paint'' as its origin is the Ge’ez word \foreignlanguage{ethiop}{'se`ile}. However, in modern Amharic writings such as newspapers and magazines, the homophonic characters are commonly observed to be used interchangeably. Since our source data are from such sources we designed our guidelines in a way to reflect the misspellings and their intended spellings.

\section{Guidelines}

We set guidelines to properly annotate misspellings collected from different sources with their contextual information. 
The guidelines are as follows. If a misspelling is not a valid Amharic word, tag it as a non-word error; if a valid Amharic word is determined to be a misspelling based on its neighboring words context, tag it as a real-word error; when deriving corrections of misspellings, adhere to intended spellings of the original authors rather than the strict Amharic writing style; and tag all words that result from informal Amharic dialects as non-word errors.

\section{Data Sources}

The data sources are textual documents obtained from random samples of Amharic news articles of Deutsche Welle and Voice of America; a retyped document of \foreignlanguage{ethiop}{ya'amAre~nAne moke^sE hohEyAte .taneqeqo yAlama.sAfe ^cegerenA mafete.hE}; and errata list of the famous Amharic novel \foreignlanguage{ethiop}{feqere 'eseka maqAbere} (Engl. Love unto Crypt).

Totally 367 sentences are annotated with guidelines presented in the previous section. The annotated corpus is released freely for research purposes.

\begin{table}
\centering
\begin{tabular}{rrr}
\hline
\textbf{Edit Distance} & \textbf{Count} & \textbf{Percentage} \\
\hline
1 & 290 & 77.96\% \\
2 & 59 & 15.86\% \\
3 & 18 & 4.84\% \\
4 & 5 & 1.34\% \\
Total & 372 & 100\% \\
\hline
\end{tabular}
\caption{The edit distance of the misspellings against their corrections.}
\label{edits}
\end{table}

\section{Results}

This section presents the number of misspellings by their types, the Damerau-Levenshtein edit distance \citep{damerau1964technique} of the misspellings from their corrections, and the correlation between the misspellings and their string lengths.

Among the 372 misspellings, 287 (77.15 \%) are found to be non-word and 85 (22.85\%) are real-word spelling errors. Two of the real-word and 34 of the non-word misspellings occur twice in the documents.

Amharic being a syllabic writing system, in order to analyze the edit distance of the misspellings from their corrections, there is a need to transliterate Amharic characters into Latin-based alphabets. The transliteration is done by following the phonetic mappings
of the popular keyboard input methods, Google and Keyman. After the transliteration process, the edit distance of the misspellings was computed against their corrections as shown in Table~\ref{edits}. About 78\% and 16\% of the misspellings are one and two edit distance from their corrections, respectively. That means about 94\% of the misspellings have two or fewer edit  distances from their corrections.

The Pearson Correlation Coefficient computed between the misspellings and their string lengths is 0.061. This indicates that there is no significant linear correlation between the misspellings and their string lengths.

\section{Conclusions}

We have developed a manually annotated corpus for Amharic misspellings that can be used to evaluate spelling error detection and correction. The misspellings are categorized as non-word and real-word errors. Besides, the availability of contextual information in the corpus makes it useful in dealing with both types of spelling errors. Moreover, the great majority of the misspellings are two or fewer edit distance away from their corrections; and there is no significant linear correlation between the misspellings and their string lengths.

\bibliography{anthology,eacl2021}

\begin{thebibliography}{10}
\expandafter\ifx\csname natexlab\endcsname\relax\def\natexlab#1{#1}\fi

\bibitem[{Baba and Suzuki(2012)}]{baba-suzuki-2012-spelling}
Yukino Baba and Hisami Suzuki. 2012.
\newblock \href {https://www.aclweb.org/anthology/P12-2073} {How are spelling
  errors generated and corrected? a study of corrected and uncorrected spelling
  errors using keystroke logs}.
\newblock In \emph{Proceedings of the 50th Annual Meeting of the Association
  for Computational Linguistics (Volume 2: Short Papers)}, pages 373--377, Jeju
  Island, Korea. Association for Computational Linguistics.

\bibitem[{Cowley(1967)}]{cowley1967standardisation}
Roger Cowley. 1967.
\newblock The standardisation of amharic spelling.
\newblock \emph{Journal of Ethiopian Studies}, 5(2):1--8.

\bibitem[{Damerau(1964)}]{damerau1964technique}
Fred~J Damerau. 1964.
\newblock A technique for computer detection and correction of spelling errors.
\newblock \emph{Communications of the ACM}, 7(3):171--176.

\bibitem[{Gezmu et~al.(2018)Gezmu, N{\"u}rnberger, and
  Seyoum}]{gezmu-etal-2018-portable}
Andargachew~Mekonnen Gezmu, Andreas N{\"u}rnberger, and Binyam~Ephrem Seyoum.
  2018.
\newblock \href {https://www.aclweb.org/anthology/L18-1651} {Portable spelling
  corrector for a less-resourced language: {A}mharic}.
\newblock In \emph{Proceedings of the Eleventh International Conference on
  Language Resources and Evaluation ({LREC}-2018)}, Miyazaki, Japan. European
  Languages Resources Association (ELRA).

\bibitem[{Grudin(1983)}]{grudin1983error}
Jonathan~T Grudin. 1983.
\newblock Error patterns in novice and skilled transcription typing.
\newblock In \emph{Cognitive aspects of skilled typewriting}, pages 121--143.
  Springer.

\bibitem[{Mitton(1985)}]{mitton1985collection}
Roger Mitton. 1985.
\newblock A collection of computer-readable corpora of english spelling errors.
\newblock \emph{Cognitive Neuropsychology}, 2(3):275--279.

\bibitem[{Mitton(2010)}]{mitton2010fifty}
Roger Mitton. 2010.
\newblock Fifty years of spellchecking.
\newblock \emph{Writing Systems Research}, 2(1):1--7.

\bibitem[{Rodrigues and Rytting(2012)}]{rodrigues2012typing}
Paul Rodrigues and C~Anton Rytting. 2012.
\newblock Typing race games as a method to create spelling error corpora.
\newblock In \emph{LREC}, pages 3019--3024. Citeseer.

\bibitem[{Tachibana and Komachi(2016)}]{tachibana2016analysis}
Ryuichi Tachibana and Mamoru Komachi. 2016.
\newblock Analysis of english spelling errors in a word-typing game.
\newblock In \emph{Proceedings of the Tenth International Conference on
  Language Resources and Evaluation (LREC'16)}, pages 385--390.

\bibitem[{Tracey and Strassel(2020)}]{tracey-strassel-2020-basic}
Jennifer Tracey and Stephanie Strassel. 2020.
\newblock \href {https://www.aclweb.org/anthology/2020.sltu-1.39} {Basic
  language resources for 31 languages (plus {E}nglish): The {LORELEI}
  representative and incident language packs}.
\newblock In \emph{Proceedings of the 1st Joint Workshop on Spoken Language
  Technologies for Under-resourced languages (SLTU) and Collaboration and
  Computing for Under-Resourced Languages (CCURL)}, pages 277--284, Marseille,
  France. European Language Resources association.

\end{thebibliography}
\bibliographystyle{acl_natbib}

\end{document}